\documentclass[11pt,a4paper]{article}
\usepackage[hyperref]{emnlp2020}
\usepackage{times}
\usepackage{latexsym}

\usepackage{microtype}
\usepackage{graphicx}
\aclfinalcopy 

\title{ExCode-Mixed: Explainable Approaches towards Sentiment Analysis on Code-Mixed Data using BERT models}

\author{
Aman Priyanshu,\And
   \hspace{0.3cm}Aleti Vardhan*,\hspace{0.5cm} \And
   \hspace{0.8cm}Sudarshan Sivakumar*,\hspace{0.5cm} \\
   Manipal Institute of Technology \\
   \texttt{\{aman.priyanshu, aleti.vardhan, sudarshan.sivakumar,} \\
   \texttt{supriti.vijay, nipuna.chhabra\}@learner.manipal.edu}\And
   \hspace{0.6cm}Supriti Vijay*, \And
   Nipuna Chhabra*\hspace{0.8cm}\\
}

\date{}

\begin{document}
\maketitle
\begin{abstract}
The increasing use of social media sites in countries like India has given rise to large volumes of code-mixed data. Sentiment analysis of this data can provide integral insights into people's perspectives and opinions. Developing robust explainability techniques which explain why models make their predictions becomes essential. In this paper, we propose an adequate methodology to integrate explainable approaches into code-mixed sentiment analysis.
\end{abstract}

\section{Introduction}
\let\thefootnote\relax\footnotetext{* All authors have contributed equally to the work.} 
Code-mixing is the mixing of two or more languages in speech. English words, for instance, are often mixed into Hindi sentences to form what is colloquially known as \emph{Hinglish} \cite{bali2014borrowing}. Social media users from multilingual countries like India are seen to express themselves using code-mixed language.
Sentiment analysis is a technique which can label this data as positive, negative, or neutral. The rise in the use of deep learning models for sentiment analysis has led to more accurate predictions. However, these complex models lack transparency on model predictions as they behave like black boxes of information. Explainable AI is an evolving area of research that provides a set of methods to help humans understand and interpret a model's decisions \cite{MILLER20191}. Specifically for the task of sentiment analysis of textual data, these techniques allow us to understand how a word or phrase influences the sentiment of the text \cite{Bodria2020ExplainabilityMF}.
This paper discusses the importance and practicality of model-agnostic explainable methods for sentiment analysis of code-mixed data.

\section{Methodology and Preliminaries}

\textbf{SAIL 2017} provides a sentiment classification dataset on Hindi-English code-mixed data that has been split into training, validation, and test sets with 10055, 1256, and 1257 instances, respectively \cite{sarkar2018jukssailcodemixed2017}. Every sample has been labeled as positive, negative, or neutral in sentiment.

\textbf{XLM-RoBERTa} is a transformer-based cross-lingual language model trained on 100 different languages \cite{conneau2020unsupervised}. We fine-tuned the model — pre-trained on Hindi-English code-mixed data — on our dataset.  We feed the outputs of the last hidden layer in XLM-RoBERTa to a softmax layer to get the final probability-class distribution.

For this work we will be discussing two model-agnostic methods, LIME \cite{ribeiro2016why} and SHAP \cite{lundberg2017unified}. 

\textbf{LIME} stands for Local Interpretable Model-Agnostic Explanations and examines the model’s behavior around an instance of the dataset rather than looking at the entire dataset. LIME works based on input perturbations and their respective changes on the model output.

\textbf{SHAP} or SHapley Additive exPlanations leverages the idea of Shapley values of game theory for model feature influence scoring. Unlike LIME, SHAP can be used to interpret feature dependency on the entire model.

LIME and SHAP use the trained model to extract predictions for the original as well as perturbed sentences. LIME interprets the individual sentences locally, while SHAP interprets the model as a whole.  The explanations are extracted for the validation and test sets.  We use these explanations to visualize and quantify our experiments.

\begin{figure*}
    \centering
    \includegraphics[scale=0.24]{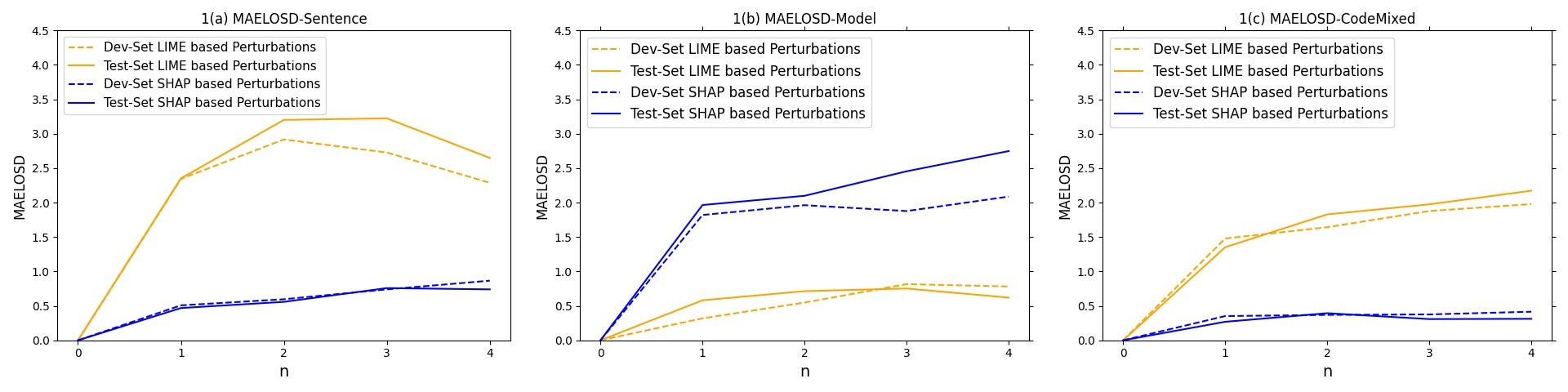}
    \caption{Metric MAELOSD for Sentence, Model, and CodeMixed is presented in this figure.}
    \label{fig:maelosd}
\end{figure*}

\section{Experimental Results}

This section discusses the results of both LIME and SHAP explainable approaches on the code-mixed data. We present a comparison between the two methodologies using three metrics, described by Mean Absolute Error of Log-Odds Scores on Deletion (MAELOSD) which is defined as,\\ 
\begin{equation}
    = \sum \frac{(|log\_odds_{i} - log\_odds_{f}|)}{n}
\end{equation}

where $i$ refers to sentences without word deletions and $f$ refers to sentences with deleted words \cite{chen2020generating}. We define polarizing words as those which have been given the highest weights by the explanations of the LIME and SHAP models.

\begin{enumerate}
  \item MAELOSD of Sentence-Interpreted Polarizing Words (\emph{MAELOSD-Sentence}): We delete the top $n$ polarizing words as returned by our explainable model from the sentence and re-compute the Log-Odds Scores. We then calculate the Mean Absolute Error (MAE) for all samples.
  \item MAELOSD of Model-Interpreted Polarizing Words (\emph{MAELOSD-Model}): We repeat the same computation with summarized weights for the entire vocabulary(English and Code-mixed). For LIME, we take the average of the word weights across each example. The new weights now represent the most polarizing words.
  \item MAELOSD of Code-Mixed Words (\emph{MAELOSD-CodeMixed}): This metric calculates sentence-wise MAE upon deletion of top $n$ polarizing code-mixed words. We consider all words which are not a part of the GloVe(6B) \cite{penningtonetal2014glove} vocabulary as code-mixed.  For instance, the word \emph{accha} is a Hindi word that does not appear in the GloVe vocabulary and is hence considered code-mixed. 
\end{enumerate}

We expect the mean absolute error to increase on the deletion of polarizing words since the model would make poor predictions. 

We can observe the results of all three metrics in Fig~\ref{fig:maelosd}. 
Fig~\ref{fig:maelosd}(a) — MAELOSD-Sentence — illustrates LIME’s local advantage over SHAP, while Fig~\ref{fig:maelosd}(b) — MAELOSD-Model — represents SHAP’s global advantage over LIME. We also observe that upon deleting $n\geq3$ words, sentences become too short, thus returning random predictions.

We can also see the impact of code-mixed data from our results in Fig~\ref{fig:maelosd}(c), where the error increases in proportion to the number of words deleted. These words have a similar impact on label prediction to that of English words. Even so, global SHAP explanations do not hold as much impact as local LIME explanations, thus implying that the \emph{Hinglish} vocabulary is much more diverse. This may be due to the deletion of globally important words that may not be present locally.

Our results align with the application of both LIME and SHAP, demonstrating their local and global natures. We can see the application and easy integration of model-agnostic interpretability pipelines on code-mixed data.

\section{Conclusion and Future Work}

Code-mixed data is an integral part of communication in multilingual communities. The use of SHAP and LIME, which quantify global and local model explanations, allows us to display their application and importance for sentiment analysis on code-mixed data.
For future work, we would like to formalize and derive a quantifiable metric for global explanations, comparing the impact of code-mixed data with standard English sentences. We would also like to experiment with different model architectures and datasets. We believe that our work serves as a valuable resource for the code-mixed AI  community. The integration of explainable methods in code-mixed data paves a new path towards future research and development.

\section*{Acknowledgments}

The authors of the paper are grateful to the reviewers in reviewing the manuscript and their valuable inputs are appreciated. We would also like
to thank the Research Society MIT for supporting the project.
\bibliographystyle{acl_natbib}
\bibliography{anthology,emnlp2020}
\end{document}